\documentclass[letterpaper]{article} % DO NOT CHANGE THIS
\usepackage{aaai2026}  % DO NOT CHANGE THIS
\usepackage{times}  % DO NOT CHANGE THIS
\usepackage{helvet}  % DO NOT CHANGE THIS
\usepackage{courier}  % DO NOT CHANGE THIS
\usepackage[hyphens]{url}  % DO NOT CHANGE THIS
\usepackage{graphicx} % DO NOT CHANGE THIS
\usepackage{natbib}  % DO NOT CHANGE THIS AND DO NOT ADD ANY OPTIONS TO IT
\usepackage{caption} % DO NOT CHANGE THIS AND DO NOT ADD ANY OPTIONS TO IT
\usepackage{subcaption}
\usepackage{algorithm}
\usepackage{algorithmic}
\usepackage{amssymb}

\usepackage{newfloat}
\usepackage{listings}
\usepackage{cleveref}
\usepackage{booktabs}
\usepackage{multirow} 
\usepackage{colortbl}
\newcommand{\method}{ReconVLA}
\definecolor{myred}{RGB}{220,60,80}     % 深亮红色（鲜艳，但不刺眼）
\definecolor{mygreen}{RGB}{6,178,240}     % 深亮绿色（标准绿色但略微加深）
\definecolor{myblue}{RGB}{96, 151, 198}      % colors
\DeclareCaptionStyle{ruled}{labelfont=normalfont,labelsep=colon,strut=off} % DO NOT CHANGE THIS
\floatstyle{ruled}
\newfloat{listing}{tb}{lst}{}
\floatname{listing}{Listing}
\nocopyright
\title{\method: Reconstructive Vision-Language-Action \\Model as Effective Robot Perceiver}
\author{
Wenxuan Song\textsuperscript{1}, 
Ziyang Zhou\textsuperscript{1}, 
Han Zhao\textsuperscript{2,3}, 
Jiayi Chen\textsuperscript{1}, 
Pengxiang Ding\textsuperscript{2,3}, 
\\
Haodong Yan\textsuperscript{1}, 
Yuxin Huang\textsuperscript{1}, 
Feilong Tang\textsuperscript{4}, Donglin Wang\textsuperscript{2}, Haoang Li\textsuperscript{1}
}
\affiliations{
    \textsuperscript{\rm 1}The Hong Kong University of Science and Technology (Guangzhou)\\
    \textsuperscript{\rm 2}Westlake University
    \textsuperscript{\rm 3}Zhejiang University
    \textsuperscript{\rm 4}Monash University
}

\usepackage{bibentry}
\usepackage{bm}
\begin{document}

\maketitle

\begin{abstract}
Recent advances in Vision-Language-Action (VLA) models have enabled robotic agents to integrate multimodal understanding with action execution.
However, our empirical analysis reveals that current VLAs struggle to allocate visual attention to target regions.
Instead, visual attention is always dispersed.
To guide the visual attention grounding on the correct target, we propose \textbf{\method}, a reconstructive VLA model with an implicit grounding paradigm. 
Conditioned on the model's visual outputs, a diffusion transformer aims to reconstruct the gaze region of the image, which corresponds to the target manipulated objects.
This process prompts the VLA model to learn fine-grained representations and accurately allocate visual attention, thus effectively leveraging task-specific visual information and conducting precise manipulation.
Moreover, we curate a large-scale pretraining dataset comprising over 100k trajectories and 2 million data samples from open-source robotic datasets, further boosting the model’s generalization in visual reconstruction.
Extensive experiments in simulation and the real world demonstrate the superiority of our implicit grounding method, showcasing its capabilities of precise manipulation and generalization.
Our project page is \url{https://zionchow.github.io/ReconVLA/}.
\end{abstract}

% Uncomment the following to link to your code, datasets, an extended version or similar.
% You must keep this block between (not within) the abstract and the main body of the paper.
% \begin{links}
%     \link{Code}{https://aaai.org/example/code}
%     \link{Datasets}{https://aaai.org/example/datasets}
%     \link{Extended version}{https://aaai.org/example/extended-version}
% \end{links}

\section{Introduction}
\label{sec:intro}

Recent progress in Vision-Language Models (VLMs) \cite{awadalla2023openflamingo, llava} has demonstrated their potential to bridge perceptual and linguistic modalities effectively. 
Building upon these advances, Vision-Language-Action (VLA) models \cite{rt1,rt2, octo_2023,niu2024llarva,song2024germ,kim2024openvla} have extended this capability to action execution by integrating multimodal understanding.
Benefit of billions of parameters and pretraining on large-scale robot datasets \cite{o2024open, fang2024rh20t}, these models have shown promise in enabling generalizable skills.

%目前，杂乱场景中和长程的manipulation任务提出了新的困难，模型需要精准的定位和切换正确的目标物体。
%精准的视觉grounding能力对于vla实现准确抓取来说至关重要，特别是在cluttered environments and long-horizon tasks中。
Accurate visual grounding is fundamental to enable precise grasping of VLAs, especially in cluttered environments and long-horizon tasks.
% present new challenges, as the model must accurately ground and switch between correct targets.
%然而，考虑下面的情况：模糊的语言、混乱的场景、需要与多个物体交互的长程任务。在这些情况下，traditional VLA 模型的视觉注意力往往表现较为分散~\Cref{}，无法聚焦于目标物体，这可能进一步导致manipulate错误的物体。这引出了一个关键问题：如何更精确的引导模型将视觉注意力聚焦于目标被操纵物体？
To analyze the visual grounding behavior during predicting actions, we visualize the attention map on visual inputs.
The results show that traditional VLA models often exhibit dispersed visual attention~(\Cref{fig: attention_vis} Row 1), failing to focus precisely on the target object, which may further lead to manipulating incorrect objects.
% Moreover, the attention allocation is even worse on long-horizon tasks with cluttered scenes that require interacting with multiple objects. 
%我们进一步进行的定量试验表明另一个造成不准确的操纵的原因来源于不充足的视觉注意力，即视觉token得到的注意力往往远低于language token的。
The finding raises a critical question: \textit{how can VLA models refine visual attention allocation and further improve visual grounding capabilities?}
% focus visual attention on the target manipulated object and guide precise manipulation?}

\begin{figure}[t!]
    \centering
    \includegraphics[width=0.48\textwidth]{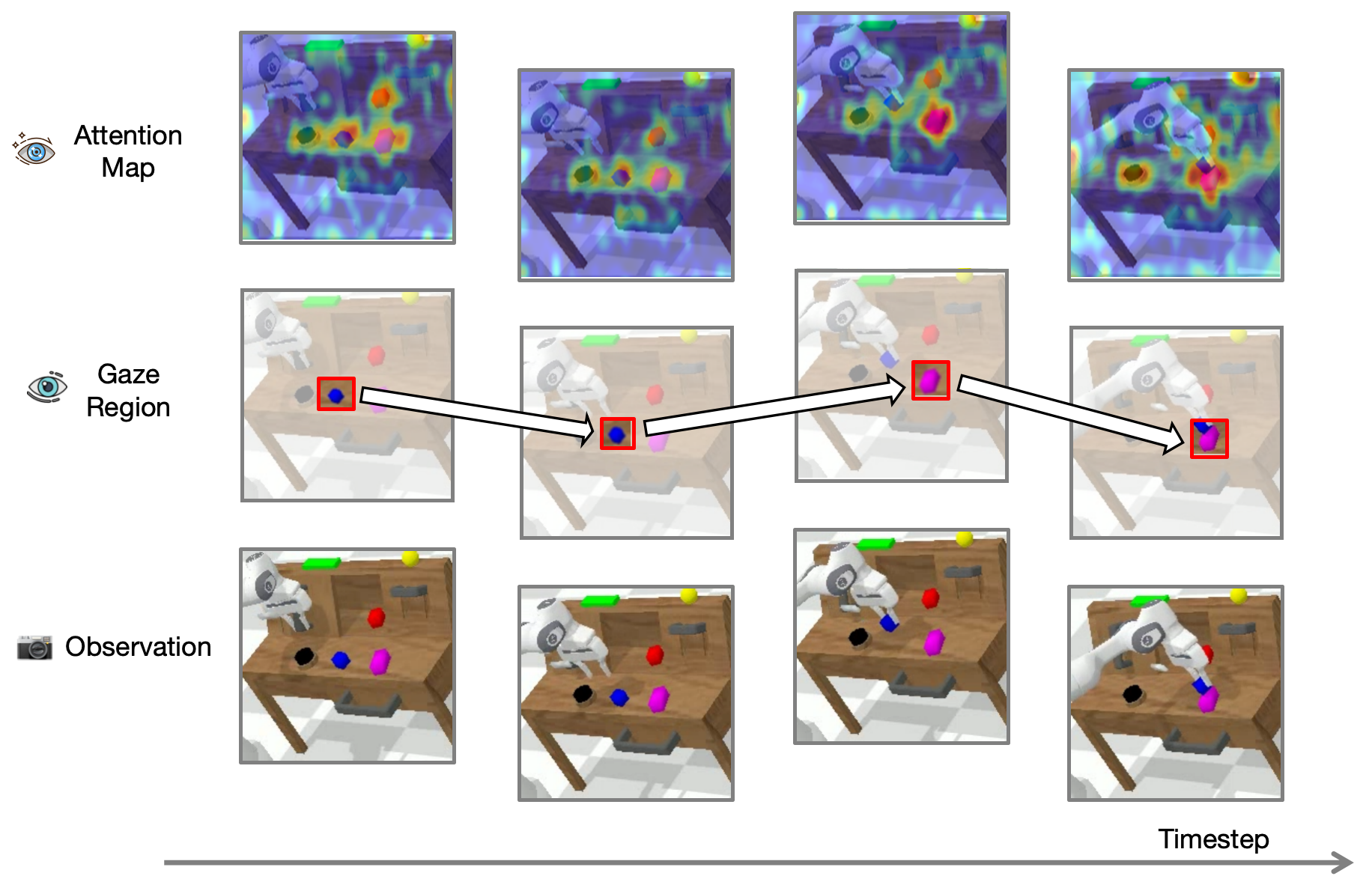}
    \caption{
    % \textbf{An example of long-horizon tasks.} For a messy desktop with an ambiguous instruction ``\textit{stack blocks}'', the robotic arm needs to first grasp one block, then stack on another.
    % Our model adaptively adjusts the gaze region, guiding the allocation of visual attention. 
    % With the precise visual grounding, it sequentially manipulates different target objects and successfully completes the task.
    \textbf{Visualization of the observation, gaze region, and attention map.}
    For a long-horizon task ``\textit{stack blocks}'' that requires the arm to lift the blue block and put it on the pink one.
    Although there are several distractors, our model adaptively adjusts the gaze region, guiding the allocation of visual attention to the right target. 
    With the precise visual grounding, it sequentially manipulates different target objects and successfully completes the task.
    }
    \label{fig:teaser}
\end{figure}

Previous visual grounding methods for VLAs usually explicitly input grounded images~\cite{roboground,virt} or output bounding boxes~\cite{ecot, graspvla} in a chain-of-thought (CoT) manner.
These methods enhance the perception of target regions and improve spatial awareness, while they do not fundamentally refine the attention allocation.
%受到ROSS的启发，我们引入了一个额外的视觉重建模块，他是一个轻量的dit模型。这个module以VLA模型的视觉输出作为条件，目标是从噪声中重建出目标操纵区域。这个过程引导VLA模型输出带有目标区域信息的表征，从而将视觉注意力聚集于正确的区域。这个过程类似于人眼的gaze动作，where the eye perceives a small, focused area with sharp clarity while the surrounding regions remain blurred~\cite{}.
Inspired by reconstructive visual instruction tuning~\cite{ROSS}, we introduce an auxiliary visual reconstruction module implemented as a lightweight diffusion transformer~\cite{Peebles2022DiT}.
This module is conditioned on the visual outputs of the VLA model and aims to reconstruct the target manipulated region from noise.
This process prompts the VLA model to learn fine-grained representations with region-specific information, thereby focusing visual attention on the correct area.
As shown in \Cref{fig:teaser}, this mechanism is analogous to the gaze behavior of the human eye, where the eye perceives a small, focused area with sharp clarity while the surrounding regions remain blurred~\cite{vision}.
Thus, the target manipulated region is named gaze region.

\begin{figure}[t!]
    \centering
    \includegraphics[width=0.47\textwidth]{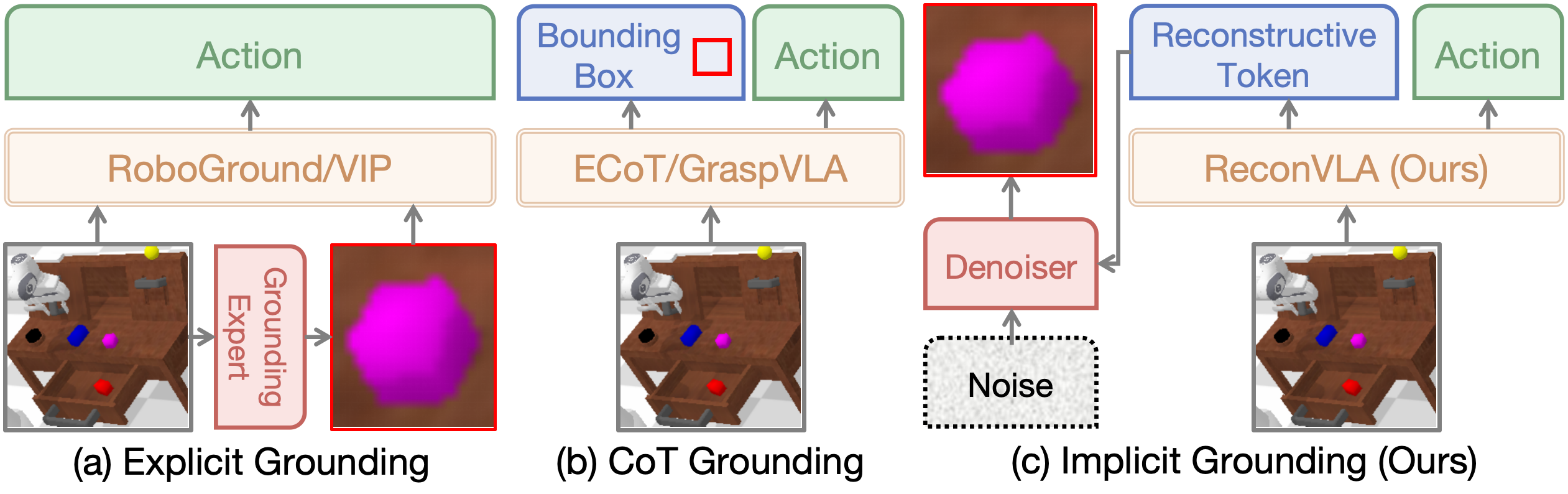}
    \caption{\textbf{Conceptual comparison between different paradigms.} \textbf{(a) Explicit Grounding}: Employing an external grounding expert and inputting entire images and cropped images~\cite{roboground,virt}.
    \textbf{(b) CoT Grounding}: Outputting coordinates of bounding boxes before action in a chain-of-thought (CoT) manner~\cite{ecot,graspvla}.
    \textbf{(c) Implicit Grounding}: Our \method~directly leverages crucial regions as implicit visual supervision for visual outputs, called reconstructive tokens, through a reconstruction process.
    }
    \label{fig:comparison}
\end{figure}

%然而，类似于their VLM backbone，常规的VLA模型只在视觉语言理解任务上进行微调，并以自回归的方式输出动作，这导致了其糟糕的视觉生成能力。因此，我们准备了一个超过100k条轨迹，2000k条数据的机器人视觉数据集。他们从开源的机器人数据集中得到，并通过Grounding Dino自动处理出pairwise的完整图像和目标区域图像。值得一提的是，由于预训练不需要机器人动作数据，我们很容易通过自动的数据处理流程将更多的互联网视频数据加入进来，从而实现scaling。在这个大规模数据集上的预训练增强了模型视觉生成的泛化性。
However, similar to their VLM backbone~\cite{llava}, conventional VLA models are finetuned on vision-language understanding tasks and generate actions in an autoregressive manner, lacking visual generation capabilities.
To address this limitation, we curated a pretraining dataset containing over 100k trajectories and 2 million data samples.
We select severak open-source robotic datasets~\cite{bridgedatav2,liu2024libero,calvin} and design an automatic data processing by Grounding DINO~\cite{groundingdino} to produce pairwise entire images and images of target manipulated regions.
% It is worth noting that since pretraining does not require robot action data, we can easily incorporate additional web-scale video data through the automated data processing pipeline, enabling scalability.
Pretraining on this large-scale dataset significantly enhances the model’s generalization ability in visual generation.

By leveraging the aforementioned techniques, we develop the \textbf{Recon}structive \textbf{V}ision-\textbf{L}anguage-\textbf{A}ction Model (\textbf{\method}).
It takes current images, language instructions, and robot proprioception as inputs. 
During training, the gaze regions of input images are processed into latent tokens via a frozen visual tokenizer, which preserves detailed visual information and enables high-fidelity reconstruction.
To better learn the latent information, we train a diffusion transformer learning to recover the latent tokens guided by reconstructive tokens.
The diffusion denosing effectively models the conditional distribution of visual observation.

Experiments in long-horizon tasks demonstrate that our implicit grounding method is more effective than other visual grounding paradigms.
Besides, visualizations of visual attention prove that our \method~demonstrates directive visual attention and leads to precise manipulation.
Then, ablation studies prove the generalization through large-scale pretraining.
Comprehensive comparison with other popular methods shows that our \method~yields superior performance.
Finally, we conduct real-world experiments and evaluate the generalization to unseen objects.
This demonstrates that our \method~has the potential to facilitate the real-world deployment of VLAs.

In summary, our key contributions are as follows:
\begin{itemize}
\item We propose \method, a reconstructive VLA model with an implicit grounding paradigm.
% The reconstruction process is a diffusion denoising process conditioned on the model's visual outputs. 
% It takes image features of the target manipulated regions as reconstructive targets to prompt VLAs toward precise visual attention allocation, thereby enhancing visual grounding capabilities and outputting precise action.
The reconstruction of gaze regions prompts the model toward precise visual attention allocation and fine-grained representation learning, thereby enhancing visual grounding capabilities and executing precise manipulation.
\item We constructed a large-scale robot pretraining dataset, containing more than 100k trajectories, 2 million data samples.
Pretraining on this dataset enhances the model's generalization of visual reconstruction capabilities.
\item Extensive experiments in simulation and the real world show the superiority of our implicit grounding methods and the capabilities of precise manipulation and generalization for unseen targets.
\end{itemize}

\section{Related Work}
\label{sec:rela}
\paragraph{Action-centric Vision-language-action Models.}
\label{sec:2.1}
Building upon foundational advancements on pretrained VLMs~\cite{beyer2024paligemma, lu2024deepseek, llava}, VLAs~\cite{rt1,rt2,quarvla,octo_2023,quartonline,more,vlas,cui2025openhelix, rationalvla, ceedvla}~learn to generate executable actions supervised by actions. 
Within them, RoboFlamingo~\cite{li2024roboflamingo} models sequential history information with an explicit policy head.
OpenVLA~\cite{kim2024openvla} is the first open-source VLA model with large-scale robotic pretraining~\cite{o2024open}.
VLAS~\cite{vlas} expands the modality with audio.
UniVLA~\cite{univla} learns task-centric latent actions from web-scale videos and adapts to different downstream tasks.
% They are typically trained on large-scale robot datasets~\cite{vima,bridgedatav2, mimicgen, droid}, which endow them with strong generalization and robustness in manipulation tasks.
These models only supervise action outputs, while our \method~supervise visual outputs as auxiliary tasks, thus enhancing visual perception.
\begin{figure*}[t!]
    \centering
    \includegraphics[width=0.85\textwidth]{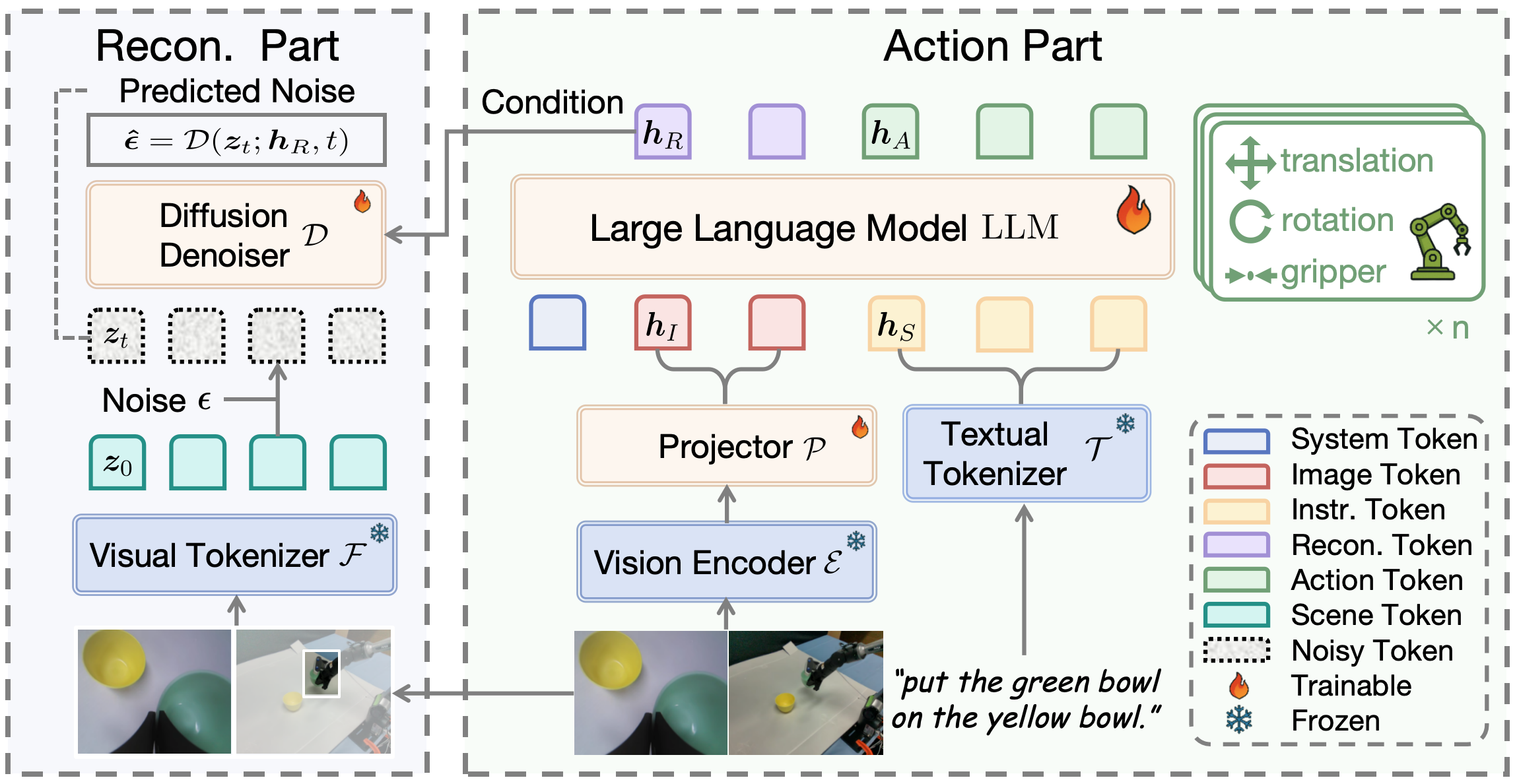}
    \caption{
    \textbf{Architecture of our \method}. 
    Our model consists of a reconstructive part and an action part.
    The input includes multi-view images and a text instruction.
    % and the reconstructive outputs are self-supervised by input images. 
    For the action part, the model outputs discrete action tokens.
    For the reconstruction part, our \method~is guided to output reconstructive tokens, which are conditions of the denoising process to reconstruct the scene tokens $z_0$ from noisy $z_t$.
    The scene tokens are tokenized images of gaze regions.
    This supervision enables our \method~to enhance visual grounding and fine-grained comprehension capabilities, which contribute to precise manipulation.
    % \textbf{Overview of our \method}. 
    % Our \method~is first pretrained on our constructed large-scale robot dataset (\textbf{A}).
    % The output consists of a reconstructive part (\textbf{B}) and an action part (\textbf{C}). 
    % % and the reconstructive outputs are self-supervised by input images. 
    % During training, \method~is guided to output reconstructive tokens, which are conditions of the denoising process to reconstruct the scene tokens $z_0$ from noisy $z_t$.
    % The scene tokens are features of the target manipulated regions of the current observation.
    % This supervision prompts \method~to accurately focus on the region to manipulate and generate high-quality actions.
    % In evaluation, our \method~demonstrates significant improvements over all baselines across both simulation benchmarks and real-world experiments (\textbf{D}).
    }
    \label{fig: arch}
\end{figure*}

\paragraph{Generative Methods for Manipulation.}
\label{sec:2.2}
Previous works have explored image or video generation models for robotic control.
Unipi~\cite{unipi} first generates future images and extracts action from generated images.
SuSIE~\cite{susie} generates subgoals with an image-editing diffusion model and executes them using a language-agnostic policy.
CLOVER~\cite{clover} produces visual plans to guide a closed-loop policy using error measurements.
GR-1~\cite{gr1} first combines generative methods with VLAs.
It proposes a GPT-style model for visual robot manipulation by leveraging large-scale video pre-training to predict future images and robot actions. 
% GR-2~\cite{gr2} scales up this approach with web-scale video pre-training, while GR-MG~\cite{gr-mg} further leverages partially annotated data and progress-guided diffusion.
3D-VLA~\cite{3dvla} further integrates depth information as guidance for vision-language-action reasoning and planning. 
GEVRM~\cite{gevrm} generates future images for a goal-conditioned policy in a closed-loop manner.
% IGOR~\cite{chen2024igor} proposes a unified latent action space by compressing visual changes between initial and goal images, enabling knowledge transfer among large-scale robot and human activity data and facilitating the training of foundation models for tasks performed by both robots and humans. 
% Seer~\cite{seer} integrates conditional visual foresight with inverse dynamics prediction in an end‑to‑end framework, attaining great robotic manipulation performance and scalability via large‑scale pre‑training. 
% PAD~\cite{guo2024prediction} proposes a diffusion‑transformer framework that jointly denoises predicted images and robot actions, enhancing robotic control performance through co‑training on diverse datasets. 
%这些方法预测未来帧的图像以从动态中学习，增强模型的planning能力，而我们的方法重建当前图像的特定区域以实现精准的感知和操纵
% Different from action-centric VLAs, these generative methods~\cite{seer, guo2024prediction, wang2025unified,worldvla} combine vision-centric tasks as auxiliary tasks.
These methods~\cite{seer, guo2024prediction, wang2025unified,worldvla} predict future frames to learn from dynamics, thereby enhancing the model's planning capability. 
In contrast, our approach reconstructs target regions of the current image to achieve precise perception and manipulation.

\paragraph{Visual Grounding Methods for Manipulation.}
\label{sec:2.3}
%显式的visual grounding往往可以作为一种视觉提示来指引模型完成正确的action。RoboGround使用LISA作为high-level 分割器，根据指令分割出正确的目标物体和背景，并将其作为观察的一部分输入给vla模型。类似的，ViRT使用YOLOv11分割出目标物体，并在放大后送入transformer-based policy。然而，这些模型借助于外部的专家模型，而没有在实质上提升policy的视觉grounding能力。EmbodiedCoT和GraspVLA通过chain-of-thought的方式依次输出Bounding Box和模型动作，在训练模型grounding能力的同时为action output提供了更多信息。区别于之前的方法，我们的模型直接从视觉输出中重建出目标区域，从而在隐式的进行grounding的同时训练了模型对目标区域的细粒度理解。这一过程模拟了人眼自发的聚焦视野内关键区域的能力。
Explicit grounding methods often take the grounded image as extra inputs to serve as auxiliary observation (\Cref{fig:comparison} (a)).
RoboGround~\cite{roboground} employs LISA~\cite{lisa} as a high-level segmenter to extract the target object and background based on instructions and feeds them as part of the observation into the VLA model.
Similarly, VIP~\cite{virt} uses YOLOv11~\cite{yolov11} to segment the target object, which is then enlarged and provided to a transformer-based policy.
However, these models rely on external expert models and do not fundamentally enhance the visual grounding capabilities of the policy itself.
ECoT~\cite{ecot} and GraspVLA~\cite{graspvla} (\Cref{fig:comparison} (b)) adopt a chain-of-thought approach, sequentially outputting bounding boxes and actions, which simultaneously trains the grounding capability and provides richer information for action output through causal attention.
In contrast to these prior methods, our \method~directly reconstructs the target manipulation region from the visual outputs (\Cref{fig:comparison} (c)), thereby implicitly performing grounding while encouraging the model to learn fine-grained representations of the target region.
This process emulates the human eye’s spontaneous ability to focus on salient regions within the field of view.

\section{Method}
\label{sec:method}
% \begin{figure}[t!]
%     \centering
%     \includegraphics[width=0.45\textwidth]{image/compare.jpg}
%     \caption{\textbf{Conceptual comparison between different pipelines.} (a) Vanilla VLAs solely leverage action supervision~\cite{rt1, rt2, kim2024openvla, li2024roboflamingo, pdvla}. (b) Generative VLAs (a kind of world models) employ the future image/video/depth as extra supervision to train the model with the planning ability~\cite{gr1, gr2, gr-mg, 3dvla, seer, unifiedvideoactionmodel, tesseract}. (c) RoboUniview~\cite{robouniview} individually trains the vision encoder with an extra 3D occupancy task to improve the perception. Both (b) and (c) rely on the large-scale pretraining. (d) Our method, in contrast, sets the input observation as extra reconstructive supervision to intrinsically activate the model to enhance the holistic comprehension capabilities and generate high-quality actions without pretraining.}
%     \label{fig:arch}
% \end{figure}

\subsection{Preliminaries}
\label{subsec:preliminaries}
To establish the foundation of our method, we first formalize the typical formulation and architecture of VLA models in the context of robotic manipulation~\cite{rt1, rt2, kim2024openvla, vlas, pdvla}.
Given a pair of images and text instructions $(I,S)$, the VLA model $\Lambda$ predicts the actions $\mathcal{A} = \Lambda(I,S)$.

\paragraph{Architecture.}
A regular VLA mainly consists of a large language model $\textnormal{LLM}$, a vision encoder $\mathcal{E}$, the tokenizer $\mathcal{T}$, and an action detokenizer~$\mathcal{Q}$.
The tuple $(I,S)$ are processed into image tokens $\bm{h}_{I}$ and text tokens $\bm{h}_S$ by $\mathcal{E}$ and $\mathcal{T}$ seperately.
These tokens are then fed into the $\textnormal{LLM}$ to generate action tokens $\bm{a}$.
Finally, the action detokenizer $\mathcal{Q}$ maps $\bm{a}$ into executable action $\mathcal{A}$ for robotic control. 
% Given the chunk size $T$ and action dimensions $D$, the length of $\bm{h}_A$ and $\mathcal{A}$ is $T \times D $.
% For single-arm manipulation, the value of $T$ is 7, which can be formulated as:
% \begin{equation}
% [X, \; Y\, \; Z, \; \phi, \; \theta, \; \psi, \; G],
% \end{equation}
% where $X,Y,Z$ represent the Cartesian coordinates of the end effector's position, $\phi,\theta,\psi$ denote the rotation angles of the end effector along each axis, and $G$ denotes the state of gripper opening.
The whole process can be formulated as:
\begin{equation}
\label{eq:1}
\mathcal{A} = {\mathcal{Q}}(\bm{a})={\mathcal{Q}}(\textnormal{LLM}(\bm{h}_{I}, \bm{h}_S)) = {\mathcal{Q}}(\textnormal{LLM}(\mathcal{E}(I),\mathcal{T}(S))).
\end{equation}
Specifically, the action tokens are generated in an autoregressive manner:
\begin{equation}
p(\bm{a})=\prod_{i=1}^N p_{\textnormal{LLM}}\left(\bm{a}_i \mid \bm{a}_{1\sim i-1} ;\bm{h}_{I};\bm{h}_S \right),
\end{equation}
where $i$ denotes the $i$-th action token and $N$ denotes the total number of action tokens.
% \paragraph{Type of Tokens.}
% Tokens in regular VLAs can be categorized into four types: system tokens, image tokens, instruction tokens, and action tokens. The system prompts usually inherit the backbone LLM and introduce the conversation background. 
% Image tokens are the latent image features extracted by a pretrained vision encoder. 
% Instruction tokens specify the manipulation task for the given image. 
% Action tokens are generated step by step and conditioned on the preceding tokens.

\subsection{Reconstructive Vision-Language-Action Model}

With observation of the dispersed attention, we aim to guide VLAs' visual attention to focus on the correct target.
Our philosophy is to construct an auxiliary visual supervision, realized by setting a reconstructive visual signal.
The supervising signal serves as conditions to guide a diffusion denoising process to reconstruct the target manipulated region.
% The visual signal comes from the target manipulated region.
Formally, we present the Reconstructive Vision-Language-Action Model (\method), grounded in this framework.

\paragraph{Reconstruction Target.}
When manipulating objects, humans receive a global view of the scene. 
However, visual perception primarily focuses on a small part of it, namely the region intended to be manipulated.
This behavior is known as gaze.
Similarly, the reconstruction target of our \method~is the target manipulated region, which we refer to as the \textbf{gaze region}.
The gaze region not only helps the model focus on the correct target among multiple affordable regions, but also enhances the detailed perception of these regions.
Besides, the mechanism implicitly facilitates sub-task planning in long-horizon tasks by focusing on and switching to different sub-goals.

\paragraph{Loss Function.}
The overall training objectives of \method~include \textbf{(i)} the autoregressive action prediction supervised by demonstration data, and \textbf{(ii)} another reconstructive term supervised by the visual features of gaze regions,
\textit{i.e.}, $\mathcal{L}_{\mathrm{\method}} = \mathcal{L}_{\mathrm{VLA}}^{\mathrm{action}} + \mathcal{L}_{\mathrm{VLA}}^{\mathrm{visual}}$, where the $\mathcal{L}_{\mathrm{VLA}}^{\mathrm{action}}$ is cross-entropy loss and the $\mathcal{L}_{\mathrm{VLA}}^{\mathrm{visual}}$ is a measurement between reconstructive tokens $\bm{h}_{R}$ and reconstruction targets $I'$.
%
%
% \begin{equation}
% \label{eq:method}
% \mathcal{L}_{\mathrm{VLA}}^{\mathrm{visual}}(\bm{h}_{R}, I) = \mathcal{M}(\mathcal{J}(\bm{h}_{R}), \mathcal{F}(I)),
% \end{equation}
%
% where $\mathcal{J}$ indicates the post projection that maps the dimensions of reconstructive tokens $\bm{h}_{R}$ to be consistent with the visual tokenizer $\mathcal{F}$. 

\paragraph{Latent Visual Reconstruction.}
% \method~performs a latent visual reconstruction.
To construct a region-specific visual supervision signal from an RGB input with spatial information redundancy~\cite{he2022masked}, we design a denosing process to reconstruct tokens with low-level features of gaze regions.
This process encourages the model to fully capture intrinsic features instead of cloning explicit RGB values~\cite{chen2023score, song2019generative, karras2022elucidating, yang2024the}.

\Cref{fig: arch} illustrates that our \method~utilizes the visual tokenizer $\mathcal{F}$ to extract target scene tokens $\bm{z}_0 = \mathcal{F}(I')$. 
Specifically, we employ a continuous variational autoencoder (VAE)~\cite{kingma2013auto} in~\cite{rombach2022high} as the visual tokenizer $\mathcal{F}$ because of its visual fidelity and ability to capture fine-grained image features.
The denoiser $\mathcal{D}$ trys to predict the noise and recover $\bm{z}_0$ from noisy tokens $\bm{z}_t$ conditioned on the reconstructive tokens $\bm{h}_R = \textnormal{LLM}(\bm{h}_I)$.
% where $\bm{h}_I=\mathcal{P}(\mathcal{E}(I))$ is the image tokens encoded by $\mathcal{E}$ and projected into LLM hidden space by a trainable projection layer $\mathcal{P}$. 
The reconstructive objective function is formalized following a diffusion process~\cite{ho2020denoising}:
\begin{equation}
    \label{eq:visual}
    \mathcal{L}_{\mathrm{VLA}}^{\mathrm{visual}} (\bm{h}_R, I') = \mathbb{E}_{t, \bm{\epsilon}} \left[
    || \mathcal{D}(\bm{z}_t; \bm{h}_R, t) - \bm{\epsilon} ||^2
    \right],
\end{equation}
where $t$ denotes the diffusion timesteps. 
The denoiser $\mathcal{D}$ consists of a stack of Transformer encoder blocks~\cite{vaswani2017attention} with self-attention modules to capture the correlations between noisy tokens and reconstructive tokens. 

%为了保证vla能够根据指令目标对应的处理视觉token，我们需要保证visual token attend to instruction tokens. 我们额外的将一组instruction token放在image token前面，以实现image token通过单向注意力得到instruction token的信息。实验证明这种interleaved格式在不影响模型固有性能的同时实现了我们的目标。

To ensure that the VLA model processes visual tokens corresponding to the instructed target, it is necessary to guarantee that image tokens attend to instruction tokens.
Thus, we prepend a set of instruction tokens before the image tokens, enabling the image tokens to fuse information from these prefix texts through causal attention.
Experimental results show that this interleaved format achieves our objective without degrading the model’s inherent language modeling capability.

\paragraph{Implementation Details.}
In this paper, we construct our \method~based on a pretrained vision-language model LLaVA-7b~\cite{llava}, which uses Qwen2-7b~\cite{qwen2} as the LLM backbone and siglip-so400m-patch14-384~\cite{siglip} as the vision encoder.

\subsection{Visual Pretraining}
%VLM backbone重建能力很弱，这是因为他是在文本和图像理解任务上训练得到。为了增强对当前观察进行重建的能力，我们设计了一个在large-scale robot dataset上针对重建任务进行预训练的过程。

%我们首先基于多个大规模的开源机器人数据集构建我们所需的预训练数据集，其中包含仿真数据和真实数据。给定其中一帧的图片和指令，我们使用微调过的最先进的grouding模型-grounding dino分割出指令中引导机械臂指向的target region，并将得到的target images和原始的holistic images存储为pair-wise的形式。通过这种方式我们得到了一个超过30k条轨迹的annotated visual pretraining数据集。
%最后，在预训练过程中我们只对重建loss进行梯度回传。
The reconstruction capability of the VLA model is inherently limited, as its VLM backbone is primarily trained on vision-language understanding tasks.
To enhance its ability to ground and reconstruct specific regions, we design a pretraining process for reconstruction tasks on a large-scale robot dataset.
% We first construct the required pretraining dataset based on several large-scale open-source robotic datasets, including both simulated and real-world data.
%预训练任务不依赖于机器人动作数据，因此原则上可以使用广泛的互联网视频数据来实现scaling。为了实现更好的域一致性，我们构建了预训练数据集based on 大规模的开源机器人数据集BridgeV2, 以及高质量的仿真数据集LIBERO和CALVIN。
% The reconstruction task does not rely on robotic action data, thus enabling the utilization of extensive internet video data for scaling. 

\paragraph{Dataset.}
% realize a generalized reconstruction with the versatility for different environments,
To build a foundational reconstruction capability, we constructed the pre-training dataset based on large-scale open-source robotic datasets BridgeData V2~\cite{bridge}, along with high-quality simulation datasets LIBERO~\cite{liu2024libero} and CALVIN~\cite{calvin}.
Given an image-text pair, we fine-tune Grounding DINO~\cite{groundingdino}, which is the state-of-the-art open-vocabulary object detector, to segment out the gaze region that the robot is instructed to interact with.
The cropped images and original images are organized in a pairwise manner.
In this way, we obtain an annotated visual pretraining dataset containing over 100k trajectories and 2 million samples.

\paragraph{Training.}
During the pretraining process, we perform gradient backpropagation both on the reconstruction loss and action loss to keep the consistency of the optimization target.
This process equips our VLM with generalized visual reconstruction capabilities and facilitates the model's deployment to diverse environments and tasks.
After pretraining, we finetune our model on specific tasks to precisely align vision-language comprehension and visual reconstruction capabilities with manipulation capabilities on the corresponding action space.

\section{Experiments}
In this section, we structure the experiments to answer the following questions: 
\begin{itemize}
    \item Does our \textbf{implicit grounding} approach outperform other visual grounding paradigms? (see \Cref{sec:4.2})
    \item Does the gazing mechanism contribute to visual grounding and further improve the precise manipulation? (see \Cref{sec:4.3})
    \item Does our proposed \textbf{pretraining} stage improve the generalization of visual generation, and how do other proposed \textbf{key designs} in \method~influence the overall performance? (see \Cref{sec:4.4})
    \item Can \method~effectively manage long-horizon tasks compared with other competitive methods? (see \Cref{sec:4.5})
    \item Can \method~realize \textbf{generalized} manipulation on \textbf{unseen} targets in \textbf{real-world} tasks? (see \Cref{sec:4.6})
\end{itemize}

\begin{figure*}[!t]
    \centering
    \includegraphics[width=0.995\textwidth]{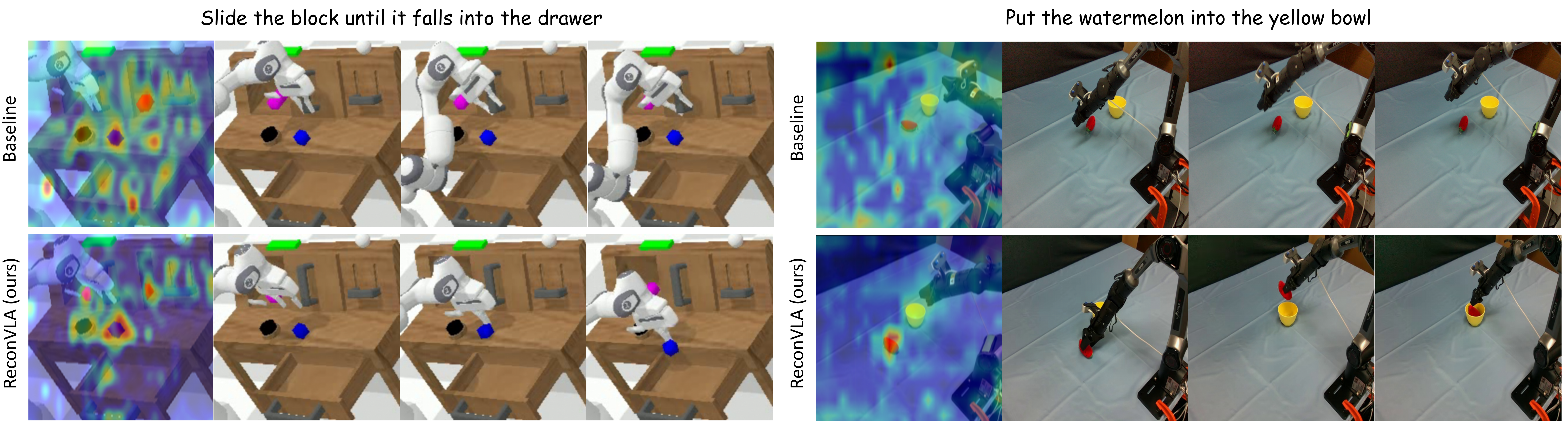}
    \caption{\textbf{Qualitative comparison of attention maps on CALVIN~\cite{calvin} and the real world.}
    \textbf{Row 1}: The baseline exhibits dispersed attention patterns or predominantly attends to an incorrect region, leading to inaccurate actions.
    \textbf{Row 2}: With auxiliary visual supervision signals, \method~forces the model to \textit{focus on specific image contents} with higher attention values and precisely move to the target region, thus successfully completing the task.
    }
    \label{fig: attention_vis}
\end{figure*}

\begin{table}[!t]
\centering
\small
% \label{tab:baselines}
\scalebox{0.99}{
\begin{tabular}{ccccccc}
\toprule
\multirow{2}{*}{Paradigm}  &
\multicolumn{5}{c}{Success Rate (\%)}  & \multirow{2}{*}{Avg. Len}\\
& 1/5 & 2/5 & 3/5& 4/5& 5/5  \\
\toprule
Baseline  & 88.8 & 76.1 & 63.7 & 57.0 & 49.0 & 3.36\\
EG  & 94.4 & 82.5 & 70.9 & 62.2 & 50.2 & 3.61\\
CG   & 47.0 & 14.3 & 1.6 & 0.0& 0.0& 0.63 \\
\midrule
IG~(ours) & \textbf{95.6} & \textbf{87.6} & \textbf{76.9} & \textbf{69.3} & \textbf{64.1}  & \textbf{3.95} \\
\bottomrule
\end{tabular}
}
\caption{Comparison among different paradigms, including Explicit Grounding (EG), CoT Grounding (CG), and our Implicit Grounding (IG). The comparison is conducted on the CALVIN ABC$\rightarrow$D.}
\label{tab:paradigm}
\end{table}

\begin{table*}[!t]
    \centering
    \small 
    \begin{tabular}{@{}cccccccccccc@{}}
    \toprule
    \multirow{2}{*}{Recon.}& Gaze& \multirow{2}{*}{Pretrain} & \multirow{2}{*}{Splits} & \multicolumn{5}{c}{Task completed in a row (\%)} & Average \\
    &Region&&& 1 & 2 & 3 & 4 & 5 & Length \\
    \midrule
    \checkmark & \checkmark & \checkmark & ABC$\rightarrow$D & \textbf{95.6} & \textbf{87.6} & \textbf{76.9} & \textbf{69.3} & \textbf{64.1}  & \textbf{3.95} \\
    \checkmark & \checkmark & $\times$ & ABC$\rightarrow$D & 96.8 & 86.9 & 76.9 & 64.9 & 58.2 & 3.85 \\
     \checkmark & $\times$ & $\times$ & ABC$\rightarrow$D & 89.8 & 80.3 & 67.7 & 56.6 & 46.5 & 3.42 \\
     $\times$ & $\times$ & $\times$ & ABC$\rightarrow$D& 88.8 & 76.1 & 63.7 & 57.0 & 49.0 & 3.36\\
    \bottomrule
    \end{tabular}
    \caption{Ablation results of the proposed techniques using the reconstructive part, gaze region, and pretraining.
    }
    \label{tab:ablation}
\end{table*}
%主线实验ABC
\begin{table*}[t]
\centering
\small
\scalebox{0.98}{
\begin{tabular}{ccccccccc}
\toprule
\multirow{2}{*}{Category} & \multirow{2}{*}{Method} & \multirow{2}{*}{Splits} &
\multicolumn{5}{c}{Success Rate (\%)}  & \multirow{2}{*}{Avg. Len}\\
& & &  1/5 & 2/5 & 3/5& 4/5& 5/5  \\
\midrule
\multirow{5}{*}{Generative Methods} 
& UniPi~\cite{unipi}~\textcolor{gray}{{(\textit{NIPS'23})}} & ABC$\rightarrow$D  & 56.0 & 16.0 & 8.0 & 8.0 & 4.0 & 0.92\\
& SuSIE~\cite{susie}~\textcolor{gray}{{(\textit{ICLR'24})}} & ABC$\rightarrow$D  & 87.0 & 69.0 & 49.0 & 38.0 & 26.0 & 2.69\\
& GEVRM~\cite{gevrm}~\textcolor{gray}{{(\textit{ICLR'25})}} & ABC$\rightarrow$D  & 92.0& 70.0& 54.0& 41.0& 26.0& 2.83 \\
& GR-1~\cite{gr1}~\textcolor{gray}{{(\textit{ICLR'24})}} & ABC$\rightarrow$D  & 85.4& 71.2& 59.6& 49.7& 40.1& 3.06 \\
& Vidman~\cite{vidman}~\textcolor{gray}{{(\textit{NIPS'24})}} & ABC$\rightarrow$D  & 91.5 & 76.4 & 68.2 & 59.2 & 46.7& 3.42\\
& CLOVER~\cite{clover}~\textcolor{gray}{{(\textit{NIPS'24})}} & ABC$\rightarrow$D  & 96.0 & 83.5 & 70.8 & 57.5 & 45.4 & 3.53\\

% & RoboUniview~\cite{robouniview} & ABC$\rightarrow$D &  87& 69.0 & 49.0 & 38.0 & 26.0 & 2.69\\
% \midrule
% \multirow{1}{*}{Diffusion Methods} 
% & 3D Diffuser Actor~\textcolor{gray}{{(\textit{CoRL'24})}}  & ABC$\rightarrow$D  & 93.8 & 80.3 & 66.2 & 53.3 & 41.2& 3.35\\
\midrule
\multirow{4}{*}{Large VLA Models} 
& VLAS~\cite{vlas}~\textcolor{gray}{{(\textit{ICLR'25})}} & ABC$\rightarrow$D  & 87.2& 64.2& 40.9& 28.1& 19.6 & 2.40 \\
& RoboFlamingo~\cite{li2024roboflamingo}~\textcolor{gray}{{(\textit{ICLR'24})}} & ABC$\rightarrow$D  & 82.4& 61.9& 46.6& 33.1& 23.5 & 2.47 \\
& OpenVLA~\cite{kim2024openvla}~\textcolor{gray}{{(\textit{CoRL'24})}}  & ABC$\rightarrow$D & 91.3 & 77.8 & 62.0 & 52.1 & 43.5& 3.27\\
& UniVLA~\cite{univla}~\textcolor{gray}{{(\textit{RSS'25})}}  & ABC$\rightarrow$D & 95.5 & 85.8 & 75.4 & 66.9 & 56.5 & 3.80 \\
     
\midrule
\rowcolor{gray!20}Reconstructive Methods & \textbf{\method~(ours)}& ABC$\rightarrow$D & \textbf{95.6} & \textbf{87.6} & \textbf{76.9} & \textbf{69.3} & \textbf{64.1}  & \textbf{3.95} \\
% \rowcolor{gray!20}Reconstructive Methods & \textbf{\method~(ours)} & ABC$\rightarrow$D & \textbf{96.8} & \textbf{86.9} & \textbf{76.9} & \textbf{64.9} & \textbf{58.2} & \textbf{3.85} \\
\bottomrule
\end{tabular}
}
\caption{\textbf{Comparison with various manipulation models on CALVIN ABC$\rightarrow$D} in success rates and average length.} 
\label{tab:baselines_ABC}
\end{table*}

%主线实验ABCD
\begin{table*}[!t]
\centering
\small
\scalebox{0.98}{
\begin{tabular}{ccccccccc}
\toprule
\multirow{2}{*}{Category} & \multirow{2}{*}{Method} & \multirow{2}{*}{Splits} &
\multicolumn{5}{c}{Success Rate (\%)}  & \multirow{2}{*}{Avg. Len}\\
& & &  1/5 & 2/5 & 3/5& 4/5& 5/5  \\

\toprule
\multirow{2}{*}{Generative Methods} 
& 3D-VLA~\cite{3dvla}~\textcolor{gray}{{(\textit{ICML'24})}} & ABCD$\rightarrow$D & 44.7 & 16.3 & 8.1 & 1.6 & 0 & 0.70\\
& GR-1~\cite{gr1}~\textcolor{gray}{{(\textit{ICLR'24})}} & ABCD$\rightarrow$D  & 94.9& 89.6& 84.4& 78.9& 73.1& 4.21 \\

\midrule
\multirow{2}{*}{Large VLA Models} 
& VLAS~\cite{vlas}~\textcolor{gray}{{(\textit{ICLR'25})}} & ABCD$\rightarrow$D  & 94.2 & 84.0 & 73.2 & 64.3 & 54.6& 3.70\\
& RoboFlamingo~\cite{li2024roboflamingo}~\textcolor{gray}{{(\textit{ICLR'24})}} & ABCD$\rightarrow$D  & 96.4& 89.6& 82.4& 74.0& 66.0 & 4.08 \\

\midrule
\rowcolor{gray!20}Reconstructive Methods & \textbf{\method~(ours)} & ABCD$\rightarrow$D & \textbf{98.0} & \textbf{90.0} & \textbf{84.5} & \textbf{78.5} & \textbf{70.5}  & \textbf{4.23} \\
\bottomrule
\end{tabular}
}
\caption{\textbf{Comparison with various manipulation models on CALVIN ABCD$\rightarrow$D} in  success rates and average length.}
\label{tab:baselines_ABCD}
\end{table*}

\subsection{Simulation Environment}
\label{sec:5.1}
% \paragraph{Simulation Environment and Experimental Details.}
The CALVIN benchmark~\cite{calvin} is built on top of the PyBullet~\cite{pybullet} simulator and involves a Franka Panda Robot arm that manipulates the scene. 
CALVIN consists of 34 tasks and 4 different environments (A, B, C and D). 
% We evaluate all methods on the classic CALVIN ABCD$\rightarrow$D setup~\cite{calvin}. 
The CALVIN long-horizon challenge is a sequential task comprising five subtasks. 
We report the success rates for each subtask and the average completed length across all five tasks. 
The method is evaluated over 500 rollouts to ensure a fair comparison.
The metrics of CALVIN are the success rates of each sub-task and the average length of all sequential 5 sub-tasks.

% \paragraph{Experiment details.}
% Our method is trained on 8 NVIDIA H100 GPUs for 2 epochs, which requires approximately 30 hours. 
% % We report the success rate and the average number of completed sequential tasks.
% The CALVIN long-horizon challenge is a sequential task comprising five subtasks. 
% We report the success rates for each subtask and the average completed length across all five tasks. 
% Each methods are tested 500 rollouts to ensure a fair comparison.

%我们将我们的方法与经典的模仿学习算法MCIL, HULC，经典的小型的VLA模型RT-1，最新的大型VLA模型VLAS，PD-VLA，以及生成式VLA模型3D-VLA进行比较。在多个子任务上的成功率以及总的平均长度现实，our \method在朴素的foundation model上实现了competative的性能，在连续的五个任务中可以完成3.73个，并且第一个任务的成功率高达94.4。

\subsection{Paradigm Comparison}
\label{sec:4.2}
We implement different visual grounding paradigms in \Cref{fig:comparison} on the same baseline to conduct a fair comparison.

\paragraph{Explicit Grounding (EG).}
We choose a finetuned YOLOv11~\cite{yolov11} as the detector to recognize the target object at each timestep.
Then we crop out the recognized object region from the image and resize it. 
Then the resized and original images are jointly fed to the VLA model to guide object manipulation. 

\paragraph{Chain-of-Thought Grounding (CG).}
For data preparation, we preprocess the images with the detector to get the coordinates of the bounding box.
Then, we reformulate the training dataset and modify the outputs in a CoT format: Bbox [x1 x2 y1 y2] + action sequence.
The input remains the original images.
In this way, the VLA model learns to ground the target object and output actions with the grounding information.

\paragraph{Results.}
As shown in \Cref{tab:paradigm}, EG gets relatively higher success rates than baseline.
%这是由于同时输入原始图片和cropped图片产生了视觉信息冗余，这无法帮助VLA无法更好的理解空间关系，反而损害了模型性能。此外，VLM backbone往往在单图上进行预训练，因此理解多图的能力较弱。
%CG的性能也较差。这表明坐标形式的bounding box不足以有效的引导模型精确manipulate正确位置。同时，直接输出坐标对VLA模型的训练是有难度的。
This indicates that explicit grounding as input helps better comprehension of spatial relationships.
%同时输出精准的坐标和动作数值给自回归vla的训练过程提出了挑战。
However, the simple concatenation of entire and cropped images introduces visual information redundancy, which limits model performance. 
CG performance is even worse. 
This suggests that bounding boxes in coordinate form are insufficient to effectively guide the model in precisely manipulating target locations. 
Additionally, directly outputting precise coordinates and action values together presents training challenges for VLA models.
%这表明我们的方法优于其他范式。这得益于我们的方法通过implicit grounding learning使得模型可以精准聚焦目标位置的视觉信息，从而实现精准操纵。此外，我们的模型直接监督视觉输出，因此不需要额外的输出和输出，从而实现了简单有效的训练和推理。

Our implicit grounding method gets the highest success rates, which demonstrates the superiority of our method over other paradigms. 
From the perspective of \textit{training mechanism}, the advantage stems from our implicit grounding learning framework, which enables the model to precisely attend to visual information at target objects, thereby achieving precise manipulation. 
From the perspective of \textit{architecture}, our model directly supervises visual outputs, eliminating the need for additional inputs or outputs. 
This design yields a simple yet effective training and inference pipeline.

\subsection{In-depth Analysis}
\label{sec:4.3}
To better explore the influence of the gazing mechanism, we conduct qualitative experiments of visual attention and its effect on fine-grained manipulation tasks.

\paragraph{Attention Visualization}
\Cref{fig: attention_vis} demonstrates that the implementation of $\mathcal{L}_{\mathrm{VLA}}^{\mathrm{visual}}$ enables the alignment of attention closely with the gaze region, which corresponds to the target object. 
For the instruction ``\textit{put the watermelon into the yellow bowl}'', the attention of baseline is highly dispersed, with the third-view image attention mostly focused on irrelevant positions and resulting in the task failure. 
In contrast, \method~successfully concentrates attention on the correct target, \textit{i.e.}, the watermelon.
This demonstrates that our method brings precise visual grounding, which facilitates task success.
% which demonstrates that precise visual grounding allows the robot to successfully open the drawer.
% Based on the above analysis, we identify the key capability: \textit{Enhanced visual attention allocation and precise visual grounding}.

\paragraph{Precise Manipulation.} 
Among all tasks, the ``\textit{stack block}'' task is the most challenging, which requires the robot to lift one block and precisely stack it on the other block.
While our baseline achieves only 59.3\% on this task, our gazing mechanism attains a success rate as high as 79.5\%, representing a \textbf{20.2\%} increase. 
This significant improvement highlights the enhanced action accuracy of our gazing mechanism through precise visual grounding.

\subsection{Ablation Study}
\label{sec:4.4}
We perform ablation studies of the proposed techniques using the reconstructive part, gaze region, and pretraining on large-scale robotic datasets in \Cref{tab:ablation}.
%我们发现预训练带来了大幅的成功率提升，这是由于对于unseen的测试环境，定位目标物体并重建难度较大，对模型生成能力的泛化性提出挑战。而大规模数据集上的预训练可以大幅度提高模型视觉生成时的泛化能力。重建要操纵的目标区域比直接重建全图更有效，这引导模型的视觉注意力集中在目标物体，避免了manipulate wrong targets。带有重建全图的模型比朴素的动作模型效果更好，这得益于对图像部分注意力的增强和更全面的理解。
We observe that \textbf{pretraining} leads to a significant improvement in success rates.
This is because, in unseen test environments, grounding the target object and performing reconstruction is inherently challenging and poses a generalization challenge to the model’s generative capability.
Pretraining on large-scale datasets substantially enhances the model’s generalization ability during visual reconstruction.
Furthermore, reconstructing the \textbf{gaze region} to be manipulated, rather than the entire image, proves to be more effective.
This guides the model’s visual attention to focus on the target object, thereby avoiding manipulation of incorrect targets.
Notably, models trained to \textbf{reconstruct} the entire image still outperform the baseline, which can be attributed to the enhanced holistic visual attention. 
However, in unseen scenarios, reconstructing the entire image with pixel redundancy is extremely challenging, which further limits the performance improvements.
% Notably, models trained to reconstruct the entire image still underperform our base model.
% This is because in unseen scenarios, reconstructing the entire image with pixel redundancy is extremely challenging, which influences the grounding performance, resulting in inaccurate actions.

\begin{figure*}[t!]
    \centering   \includegraphics[width=0.95\textwidth]{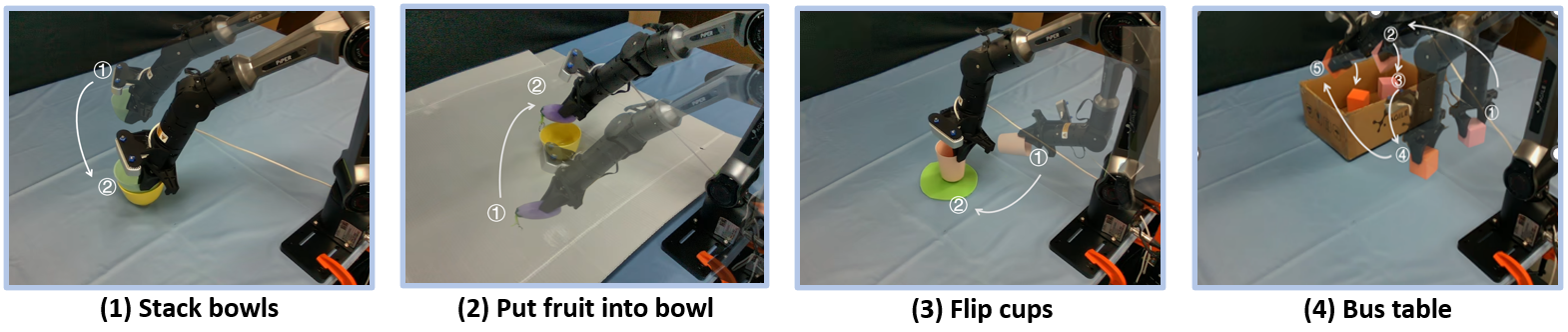}
    \caption{\textbf{Real-world Setup of four representative tasks.} 
    We use a 6-DoF AgileX PiPer robotic arm with a 1-DoF parallel gripper and a RealSense D515 depth camera as Eye-on-Base and an ORBBEC Dabai depth camera as Eye-on-Hand.  
    We selected four representative and practically meaningful tasks: (1) \textit{Stack bowls}, (2) \textit{Put fruit into bowl}, (3) \textit{Flip cups}, (4) \textit{Bus table}.}
    \label{fig:real_tasks}
\end{figure*}
\begin{figure}[t!]
    \centering
    \includegraphics[width=0.45\textwidth]{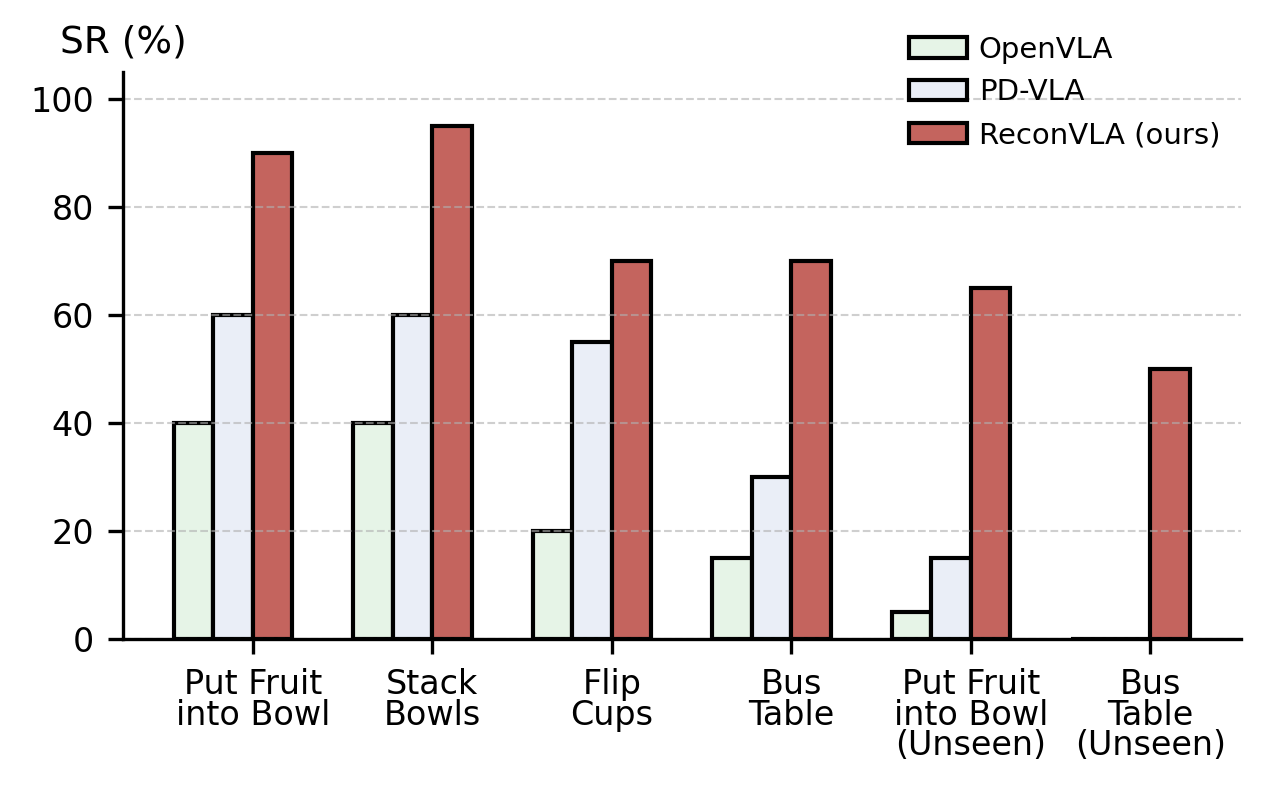}
    \caption{\textbf{Real-world multi-task results}. We report success rates (SR) across 4 different tasks as well as 2 unseen tasks.}
    \label{fig:real}
\end{figure}

\subsection{Comparison with State-of-the-arts}
\label{sec:4.5}
\paragraph{Compared Methods.}
We compare our model with generative methods that predict future images (UniPi, SuSIE, CLOVER, 3D-VLA, GR-1, Vidman, GEVRM), and large VLA models (RoboFlamingo, VLAS, OpenVLA, UniVLA), as introduced in \Cref{sec:2.2}. 
% We do \textbf{not} report metrics for those without official results.

\paragraph{Results.}
In the basic ABCD$\rightarrow$D tasks, our \method~achieves competitive performance, successfully completing an average of 4.23 out of 5 consecutive tasks, with a success rate of 98.0\% on the first task.
This indicates that our gazing mechanism provides a flexible planning ability to realize a better operation schedule in long-horizon tasks.
The ABC$\rightarrow$D tasks challenge the generalization for unseen backgrounds.
Our method surpasses all generative methods, including the popular GR-1 with over 20\% success rates on the last sub-task.
This indicates that besides the generative model that predicts the future images, enhancing the perception of the current observation is equally valuable for robot manipulation.
%在同样参数量的情况下，我们的方法也超过了openvla 20.6%以及univla 7.6%的成功率在最后一个任务上
With comparable parameter amounts, our method outperforms OpenVLA by 20.6\% and UniVLA by 7.6\% on the last sub-task, which indicates the effectiveness of our implicit grounding learning strategy.
%突破。 在所有子任务中，性能突破尤其显著的是'stack block‘这一复杂任务，相较于其他generative methods最高仅能达到58%成功率，我们的gazing mechanism在这一子任务上可以高达76%，增长18%。'stack block‘任务的动作指令抽象，操作复杂，完成难度远高于其他任务。而这一突破性的提升表明，注视机制极大增强了其在复杂任务场景下的适应性，从而显著提升了任务成功率。

\subsection{Multi-task Experiments in the Real World}
\label{sec:4.6}
%还可以加一个有遮挡的任务
% \begin{wrapfigure}{r}{0.5\linewidth}
%     \centering
%     \includegraphics[width=0.5\textwidth]{image/real_world_basic.png}
%     \caption{\textbf{Real-world multi-task results}. We report the success rate (SR) across 4 different tasks as well as average values.}
%     \label{fig:real}
% \end{wrapfigure}
\paragraph{Setup.} 
We conducted real-world experiments using a 6-DoF AgileX PiPer robotic arm with a 1-DoF parallel gripper. 
Besides, we use a RealSense D515 depth camera as Eye-on-Base and an ORBBEC Dabai depth camera as Eye-on-Hand for visual inputs.
\paragraph{Tasks.}
%我们选取了四个具有代表性和实际意义的任务: Put fruit into bowl，Put cube into bin，stack bowls，flip cups。为了增强模型的泛化能力，对于每个任务我们包含了不同的目标物体和背景颜色，每个任务的总数据量约为200条。
We select four representative tasks: \textit{Put fruit into bowl}, \textit{Stack bowls}, \textit{Flip cups}, and \textit{Bus table}. 
To enhance the model’s generalization ability, each task includes variations in target objects and background colors. 
We collect 150 trajectories per task on average.
For evaluation, each model is tested on each task with 20 trials, and the success rate is used as the performance metric.
For unseen tasks, we replace the target object with unseen ones.

% \paragraph{Experiment Details.}
%我们在4张h100上训练了我们的模型，epoch数为10，训练时长为20小时。每个任务测试20次并用成功率作为我们的指标。
\paragraph{Results.}
\method~consistently outperforms both popular OpenVLA and strong PD-VLA across the four real-world tasks, achieving the highest success rate in each case.
In particular, \method~achieves a success rate close to or exceeding 90\% on both the \textit{Put Fruit into Bowl} and \textit{Stack Bowls} tasks. 
%OpenVLA几乎无法完成flip cups和bus tables这种惊喜操纵任务，而our \method展现了较大幅度的improvements
OpenVLA shows limited effectiveness in executing fine-grained manipulation tasks (e.g., \textit{flip cup} and \textit{bus table}), while our \method~achieves significant performance improvements through precise visual grounding.
% Furthermore, it maintains strong performance with success rates above 70\% on the more challenging \textit{Flip Cups} and \textit{Bus Table} tasks.
% Compared to OpenVLA, \method~achieves a substantial absolute improvement of approximately 25\% in average success rate, highlighting the benefit of our refined attention pattern.
%在unseen场景中，由于目标物体是未在训练数据中出现过的，所有的基线都几乎丧失了成功率。得益于大规模混合数据的预训练，our \method仍然可以成功grounding目标物体并成功完成action，这体现了我们方法的视觉泛化能力带来的优势。

In unseen tasks, where the target objects are absent from the training data, both OpenVLA and PD-VLA methods exhibit nearly 0\% success rates. Benefiting from large-scale mix-data pretraining, our \method~can still successfully ground the target objects and complete the intended actions, demonstrating the advantage of our approach's \textbf{visual generalization} capability.
% These results demonstrate the robustness and generalizability of \method~in handling diverse manipulation scenarios under real-world conditions.

\section{Conclusion}
In this paper, we analyze and reveal the dispersed visual attention in traditional VLAs, which limits the precise manipulation.
% Then we propose a reconstructive vision-language-action model (\method), a novel framework that introduces reconstructive supervision via latent visual tokens.
Then, we propose a reconstructive vision-language-action model (\method), a novel framework trained in an implicit grounding paradigm.
Our \method~successfully realizes accurate visual attention allocation and further enhances manipulation skills.
We further construct a large-scale pretraining dataset for \method~to generalize on diverse scenes and unseen objects.
Extensive experiments in simulation and the real world show the superiority of our implicit grounding methods.

\bibliography{aaai2026}

\end{document}